\documentclass[letterpaper, 10 pt, conference]{ieeeconf}  
\usepackage{graphicx}
\usepackage{float} 
\usepackage[font=small]{caption}  
\usepackage{booktabs} %
\usepackage{amsmath} 
\usepackage{amssymb}  
\usepackage{cite}

\IEEEoverridecommandlockouts                              

\title{\LARGE \bf
Automatic Robot Task Planning by Integrating Large Language Model with Genetic Programming
}

\author{Azizjon Kobilov, and Jianglin Lan
\thanks{*This work was supported in part by the "El-Yurt Umidi" Foundation, and in part by the Leverhulme Trust Early Career Fellowship under Award ECF-2021-517.}
\thanks{The authors are with the James Watt School of Engineering, University of Glasgow, Glasgow, G12 8QQ, United Kingdom
        {\tt\small a.kobilov.1@research.gla.ac.uk, jianglin.lan@glasgow.ac.uk}}%
}

\begin{document}

\maketitle
\thispagestyle{empty}
\pagestyle{empty}

\begin{abstract}
Accurate task planning is critical for controlling autonomous systems, such as robots, drones, and self-driving vehicles. Behavior Trees (BTs) are considered one of the most prominent control-policy-defining frameworks in task planning, due to their modularity, flexibility, and reusability. Generating reliable and accurate BT-based control policies for robotic systems remains challenging and often requires domain expertise. In this paper, we present the LLM-GP-BT technique that leverages the Large Language Model (LLM) and Genetic Programming (GP) to automate the generation and configuration of BTs.  The LLM-GP-BT technique processes robot task commands expressed in human natural language and converts them into accurate and reliable BT-based task plans in a computationally efficient and user-friendly manner. The proposed technique is systematically developed and validated through simulation experiments, demonstrating its potential to streamline task planning for autonomous systems. 
\end{abstract}

\section{INTRODUCTION}

As technology continues to advance rapidly, the applications of autonomous systems – particularly robots – are expanding from structural environments, such as factories and industries, to unstructured and unpredictable environments, such as human houses, offices, hotels, and hospitals. In these unpredictable environments, robot control policies, particularly robot task plannings, must be created, adjusted, and updated quickly and frequently. Therefore, there is a growing demand for automated and efficient robot task planning techniques. 

Behavior Trees (BTs) have emerged as one of the most effective approaches for defining robot control policies due to their modularity, flexibility, and reusability \cite{c1,c2}. Over the years, several techniques have been developed for the automatic generation of BTs, including reinforcement learning \cite{c3}, human demonstrations \cite{c4,c5}, genetic programming (GP) \cite{c6} and others. Recently, the generation of BTs using Large Language Models (LLMs) has become one of the most active research areas, attracting significant attention from the research community \cite{c7, c8, c9, c10, c11, c12, c13, c14, c15, c16, c17, c18, c19}.

In this paper, we present the LLM and GP-based BTs Generation (LLM-GP-BT) method. In this approach, LLM is utilized to generate multiple BTs based on task descriptions written in natural human language. The LLM-generated BTs serve as the initial population for the GP evolution process. The GP evolution process refines and enhances the fitness of these BTs, ultimately producing high-fitness BTs capable of successfully executing the intended robot tasks.

Compared to the method presented in \cite{c8}, our proposed approach improves the GP process by filtering LLM-generated BTs based on their fitness levels. Additionally, our method incorporates environment-depicting images as input to provide robot-environment context, allowing users to input minimal information in the task description. The main contributions of this research are as follows:
\begin{itemize}
    \item We propose the LLM-GP-BT method to streamline the automatic generation of effective task planning for autonomous robots. We automate the environment-information input process by integrating an image analysis module. The method does not require any predefined BT examples as input, acknowledging that desired BTs are often unavailable in practical scenarios. 
    \item The proposed method uses only acceptable fitness-level BTs (generated by LLM) as initial populations, which in turn improves GP calculation process effectively. 
    \item Comprehensive simulation experiments are performed to analyze the general behavior of LLM-GP-BT and demonstrate its robustness under uncertainty and varying failure probabilities. 
\end{itemize}

\section{BACKGROUND AND RELATED WORK}

\subsection{Behavior Tree (BT)}
A BT is a graphical representation of system control used to define the task implementation policies of autonomous agents in applications such as robotics, driverless vehicles, and other autonomous systems \cite{c7}. 

A BT consists of various types of nodes categorized as root, parent, or child nodes based on their hierarchical positions \cite{c1,c2}. Figure 1 shows a simple example form of a BT containing one parent node and three child nodes. This BT can be extended by incorporating additional parent or child nodes. The complexity of a BT depends on the intricacy of the planned task. 

A BT shows the task execution structure of a particular system. The execution process begins at the root node (the parent node in Figure 1), which produces signals, so-called ticks, and sends them to its child nodes. These ticks cause child nodes to execute their associated actions, which then return a status of either Success or Failure, depending on the outcome of their execution.

A parent node, also referred to as a control flow node, can be classified into three types: Sequence, Fallback and Parallel. The behavior of ticks varies based on the type of control flow node.

-	Sequence nodes transmit ticks sequentially from left to right and return Success only if all their child nodes return Success.

-	Fallback nodes also transmit ticks sequentially from left to right but return Success if at least one of their child nodes returns Success.

-	Parallel nodes send ticks to all child nodes simultaneously and return Success if a predefined number z of child nodes return Success (where z is specified by a user).

\begin{figure}[t]
    \centering
    \includegraphics[width=0.3\textwidth]{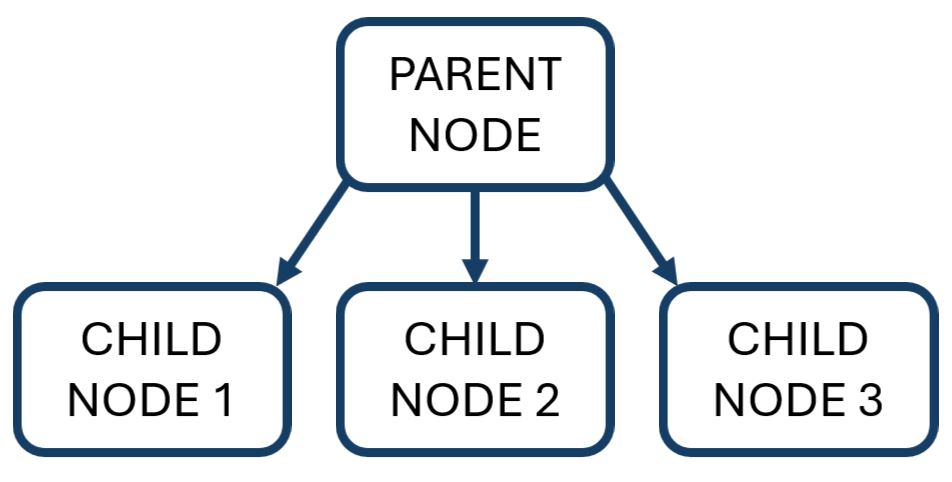} 
    \caption{Example Structure of Behavior Tree.}
    \label{fig:example}
\end{figure}

\subsection{Genetic Programming (GP)}
GP is a special algorithm that imitates natural evolution of living organisms \cite{c20}. In GP, populations of individual programs evolve generation by generation \cite{c6,c20,c21,c22}. New generations of programs are created through genetic operations such as crossover or mutation. In crossover, new offspring programs are produced by combining random parts of two individual programs. In mutation, a new program is created by randomly modifying any random part of a selected program.  Each newly generated program is evaluated using pre-defined fitness function, which assesses how well the program’s output aligns with the user’s pre-defined specifications. Only the best output programs are selected and retained for subsequent generations. This evolutionary process continues until one or more programs achieve a predefined satisfactory fitness level.

In this paper, the GP algorithm is applied to the generation of BTs. Through the iterative evolution of BTs, our approach aims to produce BTs capable of effectively executing the required tasks for an autonomous agent.

\subsection{Large Language Model (LLM)}
An LLM is a type of AI model developed to process natural language tasks such as text generation, translation, and summarization. LLMs are trained on large amounts of text data, which not only makes them proficient in text generation but also enables them to exhibit unexpected skills, known as emergent abilities \cite{c23,c24,c25,c26,c27}. These abilities include complex reasoning, decision-making, planning, instruction-following, in-context learning, and more. Interestingly, LLMs acquire these abilities despite not being explicitly trained for them. Research suggests that the emergence of such capabilities primarily results from the expansion of model size and the increasing volume and diversity of training data \cite{c23,c24,c25,c26,c27}. Today, several LLMs, such as GPT \cite{c28}, LLaMa \cite{c29}, Gemini \cite{c30}, DeepSeek \cite{c31}, and others are publicly available and widely used across various academic and research fields.

\subsection{BTs Generation using LLMs}
As LLMs continue to rapidly advance with remarkable capabilities, their integration into robotics has gained increasing attention \cite{c23,c32}. Several studies have explored the application of LLMs to robotic control, particularly in generating BT-based control policies \cite{c7}. 

Research studies \cite{c9,c10,c11} developed various techniques for the automatic generation of BTs by using LLMs, specifically GPT models \cite{c33}. In \cite{c12} researchers utilize LLMs and BTs to enable robots to explain their task execution failures. In \cite{c13}, a method for BT generation using compact LLMs was introduced, where the authors fine-tuned smaller versions of LLMs to create BTs effectively. Several studies \cite{c14,c15,c16,c17} proposed techniques that leverage LLMs to automatically expand and update BTs during robotic task planning and execution. Additionally, researchers in \cite{c18,c19} presented an LLM-based BT generation approach that incorporates the analysis of images depicting the robot's environment. 

Even though several studies have been conducted in BTs-generation field using natural language input methods, there remain open questions and gaps regarding their efficiency, accuracy and applicability to real-world scenarios. 

In \cite{c8}, the integration of LLMs with GP was proposed for BT generation. While combining LLMs with GP can be considered an effective approach, certain critical factors were not considered in their methodology. For example, their proposed framework included a validation engine to ensure the syntactic correctness of BTs generated by LLMs; however, they did not consider the functional fitness of syntactically correct BTs. Just because BTs generated by LLMs are syntactically correct does not mean they are always logically suitable for accomplishing the robot's task successfully. Furthermore, their work did not provide the statistical analysis of experimental results. Due to the phenomenon of hallucination, LLMs may generate varying BTs differently when given the same input prompt multiple times. Thus, LLMs may generate inconsistent different fitness level BTs in each usage. Additionally, they did not clarify whether examples of high-fitness BTs were included in the LLM inputs. This is an important factor, as LLMs may not always generate very precise and accurate BTs without reference examples of high-fitness BTs. In practical scenarios, however, high-fitness BTs tailored to specific robotic task planning are typically unavailable in advance. 

\begin{figure*}[t]
    \centering
    \includegraphics[width=1\textwidth]{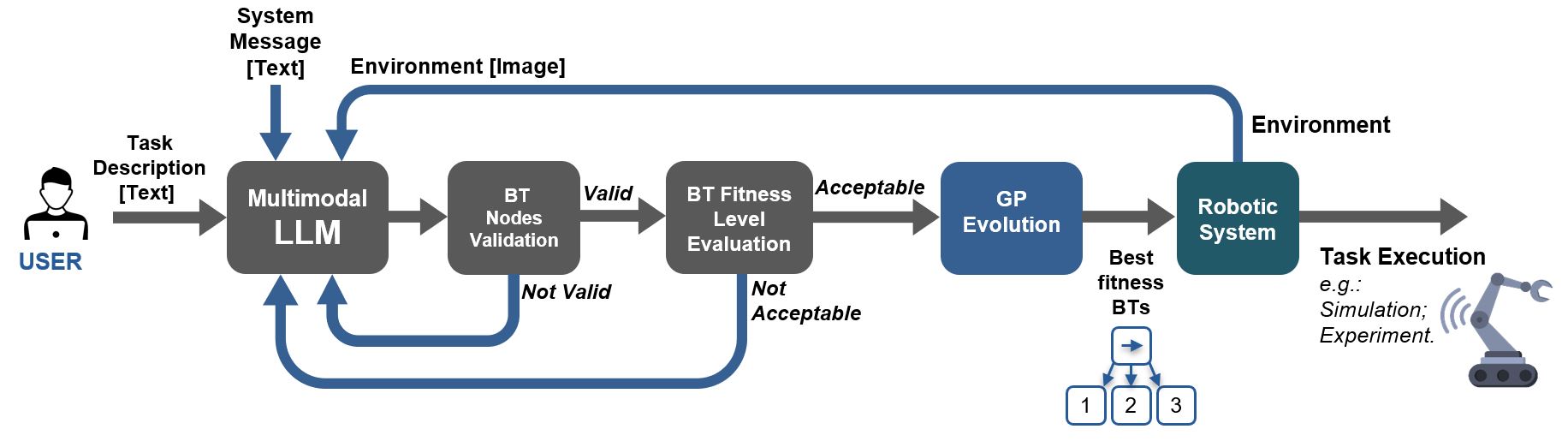} 
    \caption{LLM-GP-BT methodology framework.}
    \label{fig:example}
\end{figure*}

Our current research addresses these challenges and open questions by proposing an enhanced framework that integrates LLM with GP method for BT generation, referred to as the LLM-GP-BT technique.  

\section{METHODOLOGY}

The proposed LLM-GP-BT framework integrates an LLM and GP to generate and optimize BT-based control policies for robotic systems. Figure 2 shows the overall framework of our LLM-GP-BT methodology. We labeled LLM module as Multimodal LLM, because this unit processes both text and image domain inputs. The methodology begins by receiving three primary inputs: the system message, the robot's environment information, and task instructions.

Initially, the Multimodal LLM module is provided with a system message in natural language. The system message serves as the context that defines the role and task the LLM should adopt, outlines the expected output constraints and requirements, and specifies the available behavior functions (skills) of the robot. Essentially, the system message informs the LLM of the type of BT-based control policy it needs to generate for the system.

Additionally, the Multimodal LLM receives the robot's environmental information in image format, captured through the robot’s camera. Once the system message and environment information are provided, a user can issue task instructions to the LLM in simple natural language. For instance, in the case of Scenario-1 (discussed further in Section IV), the user might instruct: “Take the cube to the black table and place it there.” Based on these three inputs — the system message, environmental information, and task description — the Multimodal LLM generates BTs.

The LLM may occasionally produce erroneous BTs, a phenomenon referred to as "hallucination." Examples of such errors include generating BTs with unavailable robot functions, introducing unnecessary symbols or text, or producing syntactically incorrect outputs. To address these issues, a BT Nodes Validation module is employed. This module ensures that the generated BTs are contextually valid for use in the system. If the validation process identifies invalid BTs, the module requests Multimodal LLM to regenerate them until they meet the contextual requirements.

Another critical consideration is that even if the Multimodal LLM generates contextually valid BTs, these BTs may not be logically suitable for completing the assigned robotic tasks. Specifically, some generated BTs might exhibit low fitness levels, meaning they are inefficient in achieving the task goals. Low-fitness BTs negatively impact the performance of the GP process by requiring a greater number of generations until achieving the optimal solutions. To mitigate this issue, the fitness of the LLM-generated BTs must be evaluated before they are introduced into the GP process.

The fitness acceptance level can be adjusted according to the system’s accuracy requirements. A high fitness threshold may necessitate multiple iterations of BT regeneration by the Multimodal LLM. Conversely, setting the fitness threshold too low would require the GP process to perform significantly more generations to achieve the desired fitness levels of BTs. Therefore, an optimal acceptance-fitness threshold must be set to ensure that both the Multimodal LLM and GP modules operate at their optimal possible efficiency.

When the GP module is supplied with high-fitness BTs from the Multimodal LLM, it can generate optimal BTs with fewer evolution-generations. Ultimately, the resulting high-fitness BTs can be deployed effectively for robotic system control.

\begin{figure}[t]
    \centering
    \includegraphics[width=\columnwidth]{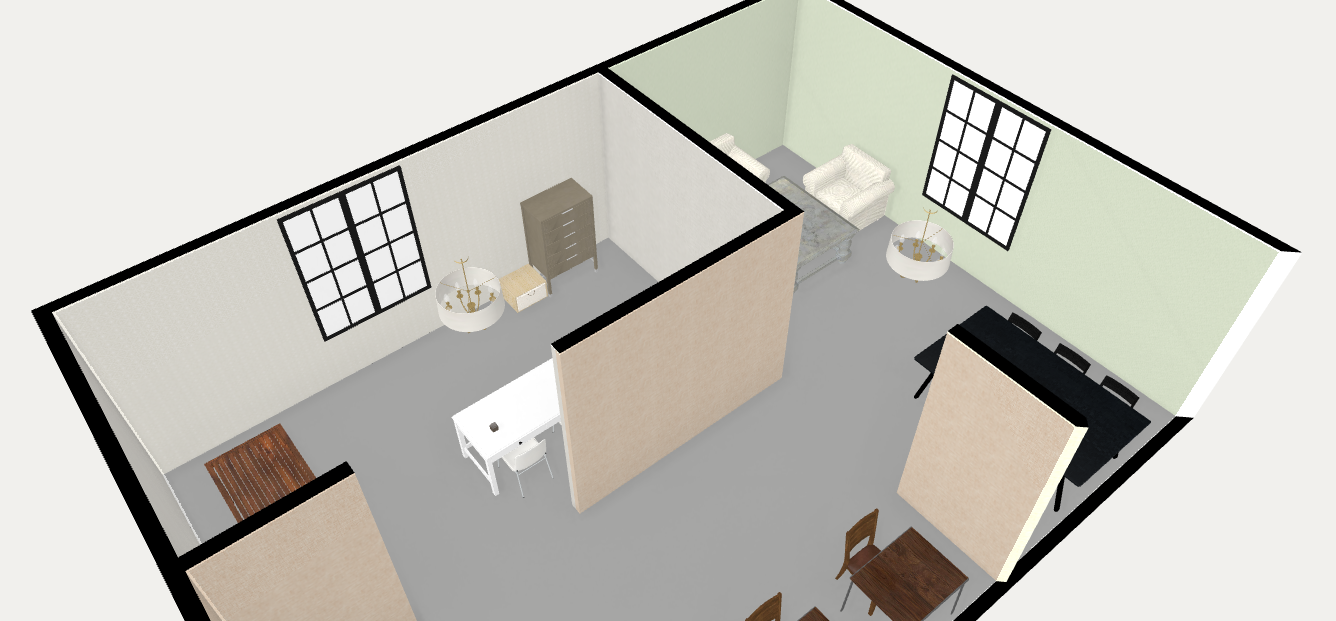} 
    \caption{Robot environment (from top perspective).}
    \label{fig:example}
\end{figure}

\begin{table}[t]
    \caption{GP Parameters and their corresponding values}
    \centering
    \begin{tabular}{|l|c|}
        \hline
        \textbf{GP Parameters} & \textbf{Values} \\ \hline
        Initial Populations & 30 \\ 
        Generations & 8000 \\ 
        Crossover \% & 40 \\ 
        Mutation \% & 60 \\ 
        Elitism \% & 10 \\ 
        Mutation: Node Mutation \% & 30 \\ 
        Mutation: Node Addition \% & 40 \\ 
        Mutation: Node Deletion \% & 30 \\ 
        Selection Method & Tournament \\ \hline
    \end{tabular}
    \label{tab:gp_parameters}
\end{table}

\section{EXPERIMENTS AND RESULTS}

\subsection{LLM-GP-BT framework setup}  
In simulations, we used GPT-4o model presented by OpanAI \cite{c3,c33} as the Multimodal LLM. To get more various BTs from LLM we adjusted two parameters of GPT model: \textit{temperature=1.2} and \textit{top}\textit{\_p=0.95}. These adjustments enable the LLM model to generate rather diverse BTs, which enhances the GP generation process. 

As for robotic system and environment, we used the same experimental setup utilized in \cite{c6}. The robot environment is depicted in Figure 3. The robot has several available functions and condition-checks to accomplish a given task. Available functions: \textit{Localise, Head up, Head down, Tuck, Pick, Place, }and\textit{ Move to TableX}. Condition-checks: \textit{Have cube? Cube X placed? }and\textit{ Task done? }

All system message information was entered into the Multimodal LLM module. For testing purposes, we designed 3D model of office environment (which is identical to the environment used in \cite{c6}), and obtained several images from that 3D model to input into our LLM module. Then, the user’s task command was entered into LLM. After that, LLM starts producing BTs for the next process, GP evolution. As for GP evolution module, we used the same GP parametric setup and fitness function applied in \cite{c6}. GP parameters used in our simulations are presented in Table 1.

In GP evolution module, the fitness function evaluates each generated BT based on its simplicity/complexity and how effectively it can complete the given task. The following function represents applied fitness function in our simulation:  
\begin{equation}\label{eq:fitness function}
J = R - \left( \alpha \| s_d - s \|^2 + \beta b + \gamma T + \delta P \right)
\end{equation}
where 
$R$ is the reward for picking the cube and placing the cube, 
$s$ is the robot state considering distance between the robot and cube, 
$s{_d}$ is the desired state, 
$b$ is the number of BT nodes,
$T$ is the task execution time, and
$P$ is the task failure probability.
$\alpha, \beta, \gamma, \delta$ are the weights presented in Table 2. $\alpha$ contains $\alpha{_1}$, $\alpha{_2}$, and $\alpha{_3}$ components. 
The same fitness function was employed in research \cite{c6,c8}.

\begin{table}[t]
    \caption{GP parameter values for function (1)}
    \centering
    
    \begin{tabular}{|l|c|}
        \hline
        \textbf{Weights} & \textbf{Values} \\ \hline
        Pick Reward & 50 \\ 
        Place Reward & 100 \\ 
        $\alpha_1$ (cube-goal distance) & 10 \\ 
        $\alpha_2$ (robot-cube distance) & 2 \\ 
        $\alpha_3$ (localization error) & 1 \\ 
        $\beta$ (BT size) & 0.5 \\ 
        $\gamma$ (execution time) & 0.1 \\ 
        $\delta$ (failure probability) & 0.0 \\ \hline
    \end{tabular} 
    \label{tab:weights_values}
\end{table}

\begin{figure}[t]
    \centering
    \includegraphics[width=\columnwidth]{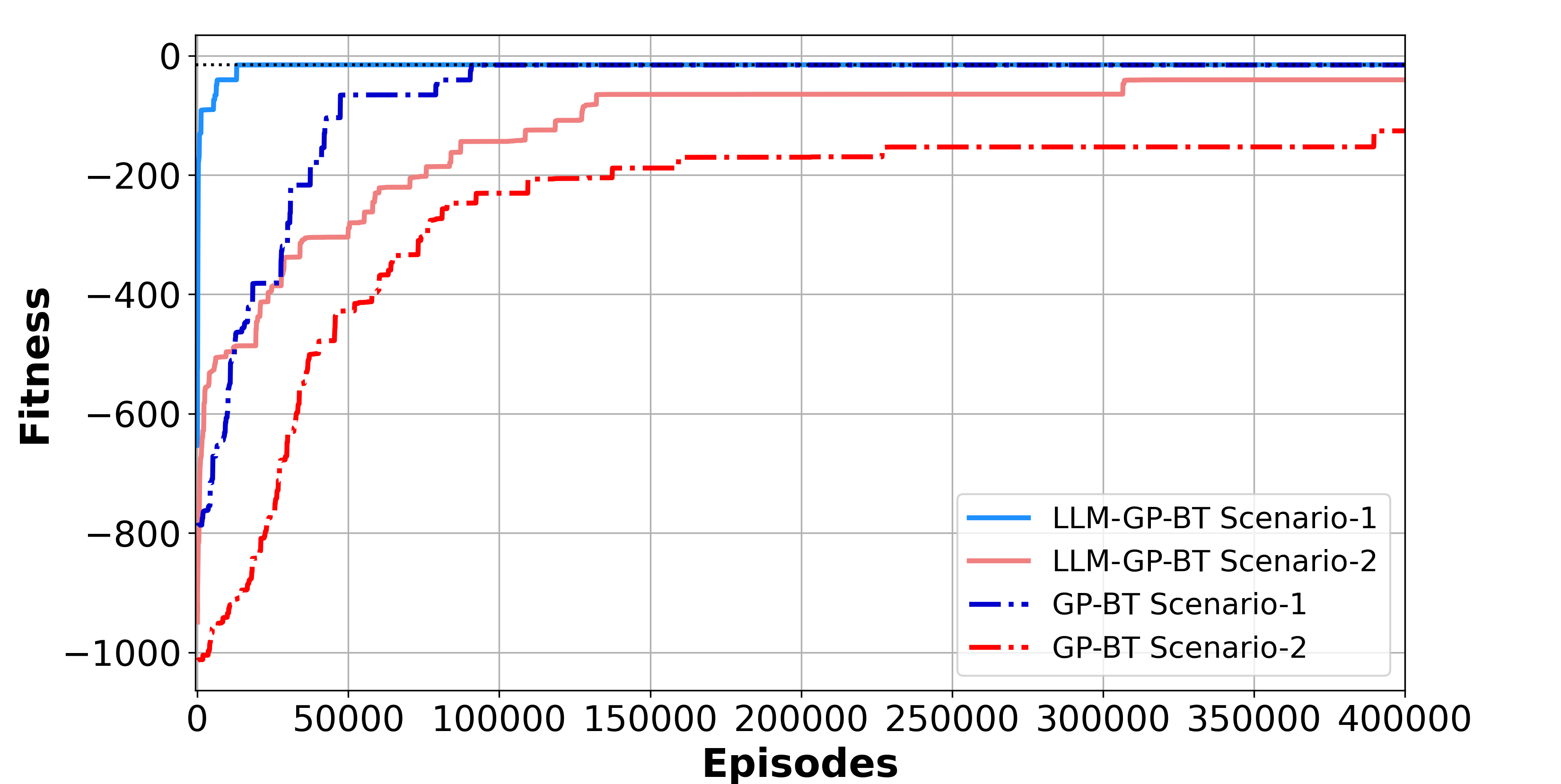} 
    \caption{BTs evolution with two different task scenarios.}
    \label{fig:example}
\end{figure}

\subsection{Experiments} 
To evaluate the efficiency of our proposed LLM-GP-BT methodology, several simulations were conducted with different scenarios and conditions. To see the improvement level of integration of LLM and GP, we also conducted simulations using pure GP-based BTs-generation method presented by \cite{c6} (which is labeled as GP-BT in this paper).  

\textbf{Experiment 1:} We tested how LLM-GP-BT can generate BTs for two different task scenarios. In Scenario-1, the robot must pick the cube from one table and place it onto another target table. In Scenario-2, the robot must pick the cube from one of the three given tables in the environment and bring it onto the target table, where the exact location of the cube is uncertain.

To statistically analyze the effectiveness of the proposed technique, we run the simulations 10 times for each scenario and presented average-mean values in the result. Figure 4 shows the BTs-fitness evolutions using LLM-GP-BT and GP-BT methods for two scenarios. As can be seen from the line chart, LLM-GP-BT method achieved the highest fitness level in a very few episode generations compared to GP-BT method. In the figure, episodes represent the number of all generated BTs during GP evolution. 

\begin{figure}[t]
    \centering
    \includegraphics[width=\columnwidth]{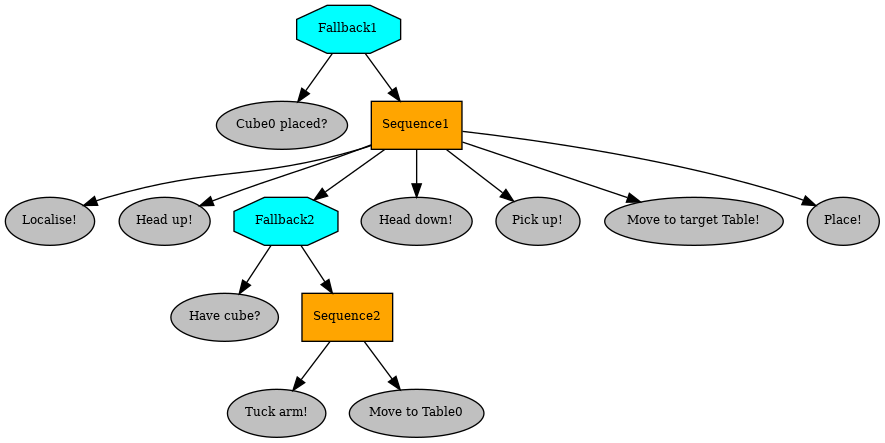} 
    \caption{LLM-GP-BT produced optimal fitness BT output for Scenario-1.}
    \label{fig:example}
\end{figure}

\begin{figure}[t]
    \centering
    \includegraphics[width=\columnwidth]{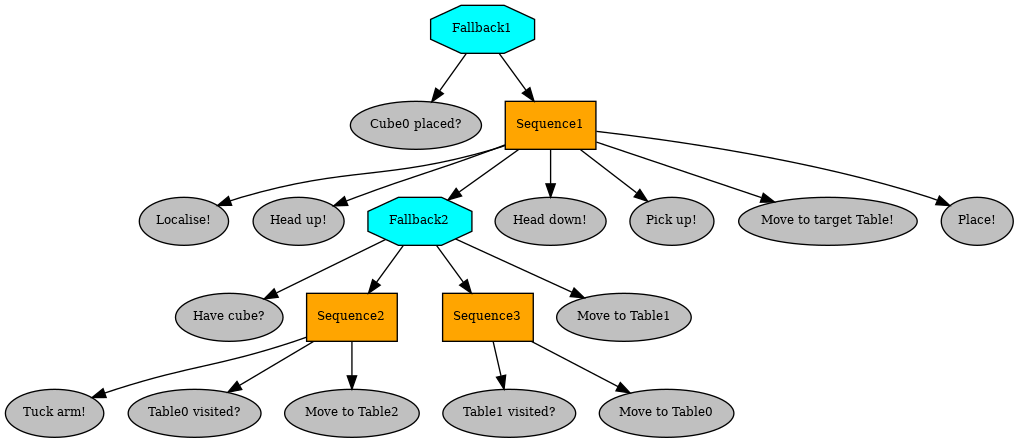} 
    \caption{LLM-GP-BT produced optimal fitness BT output for Scenario-2.}
    \label{fig:example}
\end{figure}

The samples of BTs with optimal fitness levels, obtained from simulations using the LLM-GP-BT technique, are presented in Figures 5 and 6. These BTs closely resemble the results reported in the GP-BT study \cite{c6}, as the same task, environment, and BT-fitness-evaluation function were employed in both researches.

\textbf{Experiment 2:} We tested our methodology under deterministic and stochastic conditions, where robot actions have failure probability. Table 3 presents the failure probabilities associated with each condition. 

\begin{table}[t]
    \caption{Failure probability values}
    \centering
    \begin{tabular}{|l|c|c|c|c|}
        \hline
        \textbf{Probability Type} & \textbf{Det.} & \textbf{Stoch.-1} & \textbf{Stoch.-2} & \textbf{Stoch.-3} \\ \hline
        Localization failure & 0 & 0 & 0.2 & 0.3 \\ 
        Picking failure      & 0 & 0 & 0.2 & 0.4 \\ 
        Placing failure      & 0 & 0 & 0.1 & 0.2 \\ 
        Losing cube          & 0 & 0.05 & 0.05 & 0.1 \\ 
        Losing localization  & 0 & 0.1 & 0.1 & 0.2 \\ \hline
    \end{tabular}
    \label{tab:failure_probabilities}
\end{table}

\begin{figure}[t]
    \centering
    \includegraphics[width=\columnwidth]{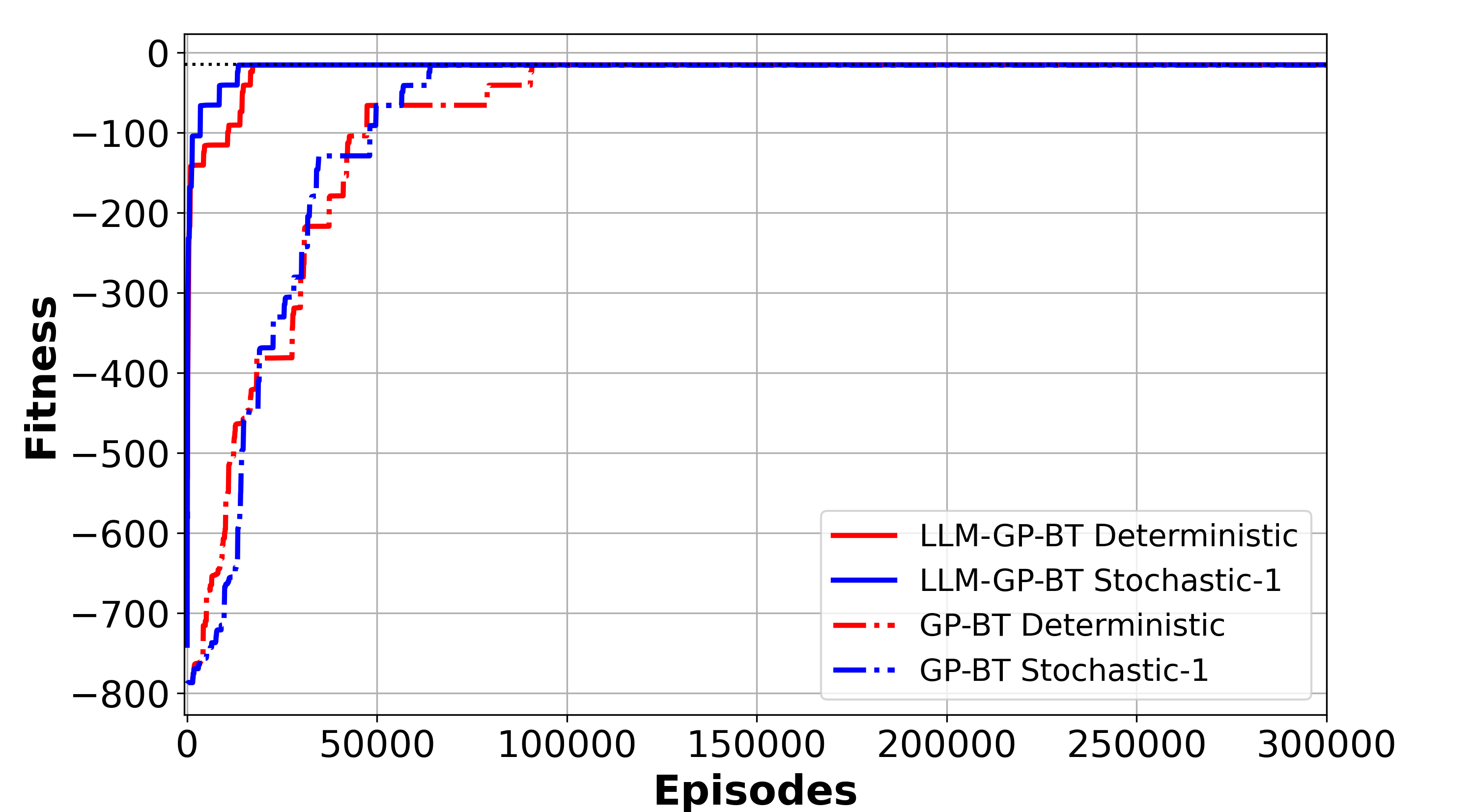} 
    \caption{BTs evolution with increasing levels of uncertainty.}
    \label{fig:example}
\end{figure}

\begin{figure}[t]
    \centering
    \includegraphics[width=\columnwidth]{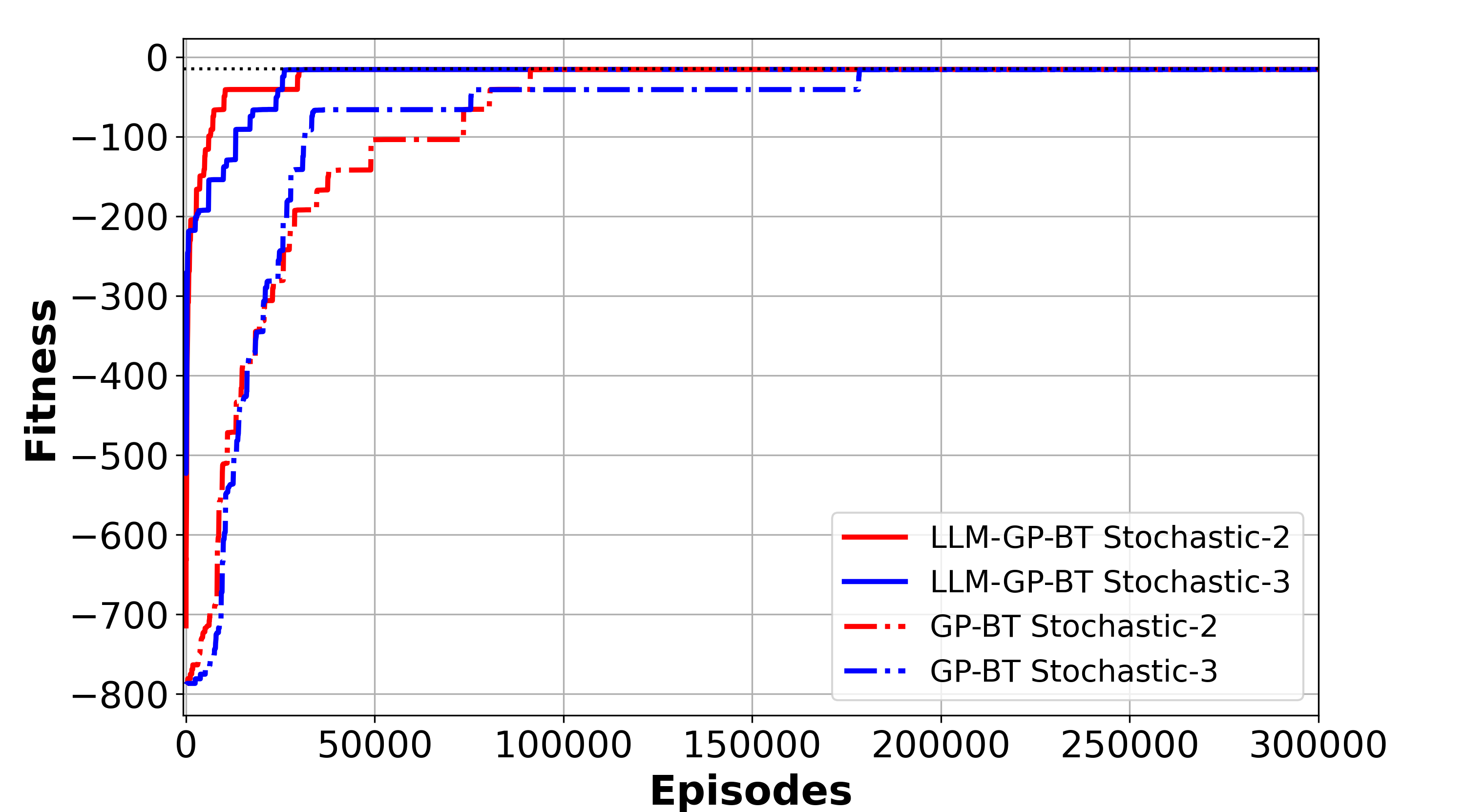} 
    \caption{BTs evolution with increasing levels of uncertainty.}
    \label{fig:example}
\end{figure}

Figures 7 and 8 depict the fitness evolution of BTs in deterministic and stochastic scenarios (with failure probabilities) for both the LLM-GP-BT and GP-BT methods. The results clearly demonstrate that our proposed approach achieved an optimal fitness level more rapidly than the GP-BT method both in deterministic and stochastic cases. These results highlight the robustness and efficiency of the LLM-GP-BT method in handling environments characterized by increasing levels of uncertainty. 

Generating a larger initial population of BTs requires more computational resources for LLM-GP-BT performance. Therefore, we also tested reducing the initial population size in our technique.

\textbf{Experiment 3:} We conducted simulations using the LLM-GP-BT methodology with reduced initial populations of 30, 20, and 15 BTs. As shown in Figure 9, decreasing the initial population size slows the rate of convergence but does not significantly affect the trend toward achieving optimal fitness. Notably, the convergence rate of LLM-GP-BT with an initial population of 15 outperformed the GP-BT method with an initial population of 30.  

It is important to note that a larger initial population typically enables the GP process to achieve optimal fitness in fewer generations. However, with the LLM-GP-BT technique, even a smaller initial population generated by the LLM allows the GP process to reach optimal fitness BTs in fewer generations. These findings indicate that the proposed LLM-GP-BT method maintains robust performance and achieves faster convergence, even with reduced initial population sizes.

\begin{figure}[t]
    \centering
    \includegraphics[width=0.5\textwidth]{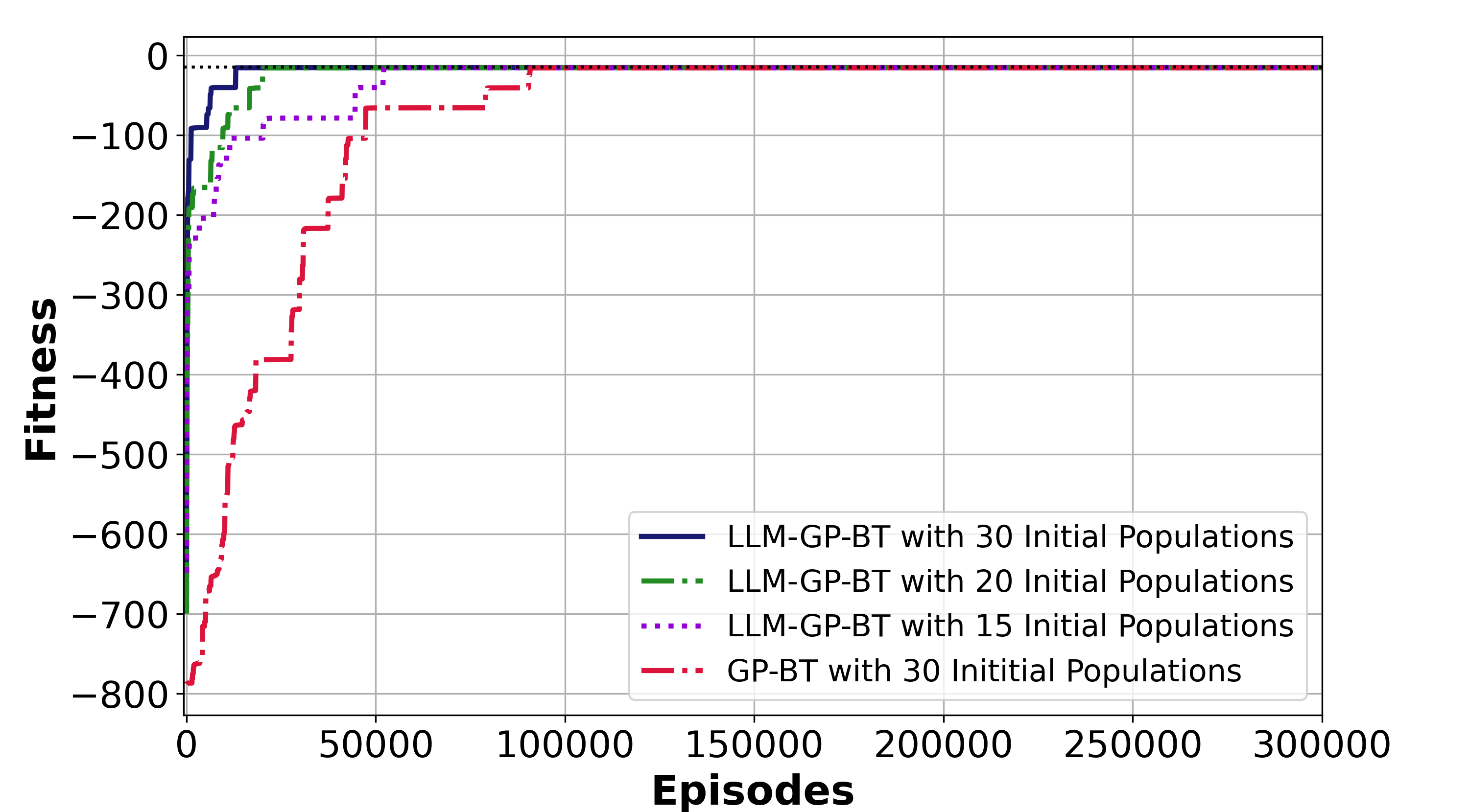} 
    \caption{BTs evolution with decreased initial populations in LLM-GP-BT.}
    \label{fig:example}
\end{figure}

\section{DISCUSSION AND FUTURE WORK}
LLMs can instantly generate BTs with optimal fitness levels if an example of a high-fitness BT is provided as part of the primary input. In such cases, the fitness evolution figure would show an immediate achievement of the optimal fitness level. However, in real-world scenarios, the desired BTs are typically not known beforehand, making it impossible to supply examples of ideal BTs in advance. To ensure our simulation reflects practical conditions, we did not provide any examples of desired BTs as inputs. Instead, we included descriptions in the system message and provided the necessary format information for output BTs.

For fair comparisons in our simulation experiments, we adopted the same GP parameter values used in a prior GP-BT study \cite{c6}. In that research, the initial populations comprised randomly generated, small-sized BTs, and the GP parameters were tuned to efficiently evolve these simple BTs. In contrast, the LLM-GP-BT technique generates more complex initial BT populations tailored to the specified tasks. As a result, further adjustment and fine-tuning of GP parameters could enhance the performance of the LLM-GP-BT method, which is an avenue worth exploring in future research.

If more detailed information about the robotic system, its tasks, and its environment is included in the system message and the user’s task command (inputs for the LLM), the LLM could instantly generate BTs with optimal fitness. This approach would allow for a reduction in both the initial population size and the number of GP generations needed to efficiently achieve the desired fitness. Conversely, if only limited or general information is available for the system message and task command, it becomes advantageous to generate a larger initial population of BTs using the LLM. This strategy ensures that the GP process can efficiently evolve BTs to achieve the highest fitness levels. Therefore, the computational efficiency of the LLM-GP-BT method depends significantly on the quality and quantity of information provided in the system message and task command.

Crafting effective prompts is crucial for maximizing the potential of LLMs. Well-designed and organized prompts can enable LLMs to generate better BTs for task planning. Hence, further investigation into prompt engineering methods and their impact on LLM-based BT generation is necessary. 

Additionally, the adjustment of LLM parameters, such as \textit{temperature} and \textit{top}\textit{\_p}, affects the diversity of the generated BTs. Higher values for these parameters tend to produce a broader range of BT variations. Exploring how these parameter adjustments influence the performance of the LLM-GP-BT approach is another promising area for future research.

\section{CONCLUSIONS}
This paper presents an integrated framework that combines LLM with GP to efficiently generate BT-based control policies for autonomous systems. The proposed method demonstrated improved efficiency by leveraging acceptable-fitness initial populations, optimizing computational resources, and incorporating robot-environment image analysis. Simulation results confirm its robustness under various conditions, including uncertainty and reduced initial population sizes. Overall, the proposed LLM-GP-BT technique offers a scalable and efficient solution for enhancing autonomous system task planning with minimal human input.

\addtolength{\textheight}{-12cm}   






\end{document}